\documentclass[AMA,Times1COL]{WileyNJDv5} %STIX1COL,STIX2COL,STIXSMALL

\articletype{Article Type}%

\received{14 Jul 2023}
\revised{25 Aug 2023}
\accepted{27 Aug 2023}
\journal{Journal}
\volume{00}
\copyyear{2023}
\startpage{1}

\newcommand{\revise}[1]{\textcolor{black}{#1}}

\begin{document}

\title{Autonomous Soft Tissue Retraction Using Demonstration-Guided Reinforcement Learning}
\titlemark{Autonomous Soft Tissue Retraction Using Demonstration-Guided Reinforcement Learning}

\authormark{Singh \textsc{et al.}}
\author[1]{Amritpal Singh}
\author[2]{Wenqi Shi}
\author[3]{May D Wang}
\address[1]{\orgdiv{College of Computing}, \orgname{Georgia Institute of Technology}, \orgaddress{\state{Georgia}, \country{USA}}}
\address[2]{\orgdiv{Electrical and Computer Engineering}, \orgname{Georgia Institute of Technology}, \orgaddress{\state{Georgia}, \country{USA}}}
\address[3]{\orgdiv{Wallace H. Coulter Department of Biomedical engineering}, \orgname{Georgia Institute of Technology}, \orgaddress{\state{Georgia}, \country{USA}}}
\corres{Dr. May D. Wang, \email{maywang@gatech.edu}}

%The Medical Image Computing and Computer Assisted Intervention Society
\publishedin{This work is accepted in MICCAI 2023 - Augmented Environments for Computer Assisted Interventions (AE-CAI) workshop}

% \author[1]{****************}
% \author[2]{****************}
% \author[3]{****************}
% \address[1]{\orgdiv{****************}, \orgname{****************}, \orgaddress{\state{****************}, \country{****************}}}
% \address[2]{\orgdiv{****************}, \orgname{****************}, \orgaddress{\state{****************}, \country{****************}}}
% \address[3]{\orgdiv{****************}, \orgname{****************}, \orgaddress{\state{****************}, \country{****************}}}
% \corres{****************, \email{****************}}

% \presentaddress{This is a sample for present address text.}

% \jnlcitation{\cname{%
% \author{Singh A.},
% \author{Shi W.}, and
% \author{wang M.}
% \ctitle{Autonomous Soft Tissue Retraction Using Demonstration-Guided
% Reinforcement Learning} \cjournal{\it J Comput Phys.} \cvol{2021;00(00):1--18}.}

% \fundingInfo{Text}
% \JELinfo{ejlje}

\abstract[Abstract]{In the context of surgery, robots can provide substantial assistance by performing small, repetitive tasks such as suturing, needle exchange, and tissue retraction, thereby enabling surgeons to concentrate on more complex aspects of the procedure. However, existing surgical task learning mainly pertains to rigid body interactions, whereas the advancement towards more sophisticated surgical robots necessitates the manipulation of soft bodies. Previous work focused on tissue phantoms for soft tissue task learning, which can be expensive and can be an entry barrier to research. Simulation environments present a safe and efficient way to learn surgical tasks before their application to actual tissue. In this study, we create a Robot Operating System (ROS)-compatible physics simulation environment with support for both rigid and soft body interactions within surgical tasks. Furthermore, we investigate the soft tissue interactions facilitated by the patient-side manipulator of the DaVinci surgical robot. Leveraging the pybullet physics engine, we simulate kinematics and establish anchor points to guide the robotic arm when manipulating soft tissue. Using demonstration-guided reinforcement learning (RL) algorithms, we investigate their performance in comparison to traditional reinforcement learning algorithms. Our in silico trials demonstrate a proof-of-concept for autonomous surgical soft tissue retraction. The results corroborate the feasibility of learning soft body manipulation through the application of reinforcement learning agents. This work lays the foundation for future research into the development and refinement of surgical robots capable of managing both rigid and soft tissue interactions. Code is available at \href{https://github.com/amritpal-001/tissue_retract}{https://github.com/amritpal-001/tissue\_retract}.}
\maketitle

\footnotetext{\textbf{Abbreviations:} DG-RL = Demonstration guided reinforcement learning; STL = Surgical task learning; PSM = Patient side manipulator; \revise{ROS = Robot operating system; RL = Reinforcement learning; MDP = Markov Decision Process; DDPG = Deep deterministic policy gradient;} }

\renewcommand\thefootnote{\fnsymbol{footnote}}
\setcounter{footnote}{1}

\section{Introduction} \label{intro}
%Clinical motivation 
    % Can robots assist humans in soft tissue surgical tasks? 
    % Can we learn this inside simulation, instead of real tissue? 

Over the past two decades, robotic surgery has revolutionized the field of surgical procedures. There is a growing literature on unmet surgical needs, the disparity in surgical access \citep{mylesGlobal2010,debasEmergence2015}, and surgeon fatigue \citep{sturm_effects_2011}. The introduction of robotic surgery, combined with a growing digital footprint, has made it increasingly feasible to assist or automate specific subtasks. Robotic surgical systems are capable of supporting surgeons in performing repetitive tasks such as suturing and handover of needles. This reduces the overall duration of surgical procedures and enhances their efficiency. Simulations with physics engines such as pybullet \citep{pybullet} and mujoco\citep{todorov_mujoco_2012} have been effectively utilized to build accurate representations of surgical robots, training algorithms, and their subsequent validation in real-world scenarios\citep{tagliabue_soft_2020,shin_autonomous_2019,dettorre_learning_2022}. 
Such simulation environments offer a risk-free platform to develop these systems, eliminating the potential for harm to both the robotic arm and tissue. 

% Rigid vs Soft body simulation env?
% Technical motivation 
    % Can we simulate soft and rigid body dynamics for surgical scenarios? 
    % Can we train RL algorithms to learn soft tissue manipulation? 
    % Can we train RL algorithms with sparse reward (and no reward engineering)?
    % Can we create an expert policy to generate guidance data? 

Most of the current research on surgical task learning is mainly confined to rigid-body interactions, such as needle grasping, gauze transfer, and needle handover \citep{xu_surrol_2021, dex}. However, real-world surgical procedures involve the continuous manipulation of the soft body, which underscores the necessity of mastering soft body manipulation to advance surgical systems 
\citep{singh_roadmap_2022}. Unlike rigid body simulation, soft bodies can change in shape or length. Though the relative distance between points will change, the soft bodies retain their shape to some degree. Unlike rigid bodies, soft bodies introduce elastic pull or recoil toward their original position. 

\revise{Murali et al.\citep{murali_learning_2015}, Attanasio et al. \citep{attanasio_autonomous_2020}, and Saeidi et al. \citep{saeidi_autonomous_2022} learn soft tissue manipulation using tissue phantoms, real tissue or cadaver. This can be expensive and can be a barrier to entry into soft tissue manipulation research. Phantom tissue fabrication can have a high cost, low temporal stability, and involve laborious methodologies for development \citep{hernandez-quintanar_discovering_2018, cabrelli_stable_2016}. As an alternative, soft tissue simulations offer a promising avenue for developing such systems prior to real-world testing. Tagliabue et al. \citep{tagliabue_soft_2020} introduced UnityFlexML, a Unity-based system, designed to interact with the patient-side manipulator (PSM) through soft body interactions. Although the use of tissue phantoms or real tissue can yield robust results, these resources can be prohibitively expensive and not always readily available.} 

\revise{Advantageously, soft tissue simulations enable precise measurement of exerted forces, allowing control of training agents with minimal or no tissue damage, a feature that can prove challenging when using tissue phantoms. With advances in physics engines, simulations can also allow for domain randomization to vary tissue color, size, location, tissue elasticity, and gravity. As such, simulations offer a valuable and scalable tool in advancing the field of surgical robotics.
}
Pioneering work in soft tissue manipulation has harnessed the capabilities of reinforcement learning algorithms \citep{tagliabue_soft_2020}. Despite their potential, reinforcement learning algorithms can encounter difficulties due to the vast exploration space of surgical tasks. The challenge here lies in the inherent complexity of the surgical environment, as well as the precise control required for such tasks. Leveraging \revise{demonstration data} may offer a viable solution for mitigating this issue, by substantially reducing the exploration space and guiding the learning process. 

 % Problem statement 
\revise{In the present study, we discuss the complexities of soft tissue interactions with DaVinci robots and explore the use of a simulation environment to learn soft tissue manipulation tasks. To achieve this, we generate a simulation environment supporting soft and rigid body interactions and train agents to perform tissue retraction tasks.} We formulate a \revise{rule based} policy to generate demonstration data to guide the training of reinforcement learning algorithms.
In summary, our primary contributions to this work are threefold:

% Our contributions 
    \begin{itemize}
        \item 
        We establish a ROS-compatible physics simulation to replicate soft and rigid interactions for the DaVinci robot, particularly tailored to the tissue retraction task. We further explore the increased complexity of soft and rigid-body tasks compared to rigid-body interactions. In this context, we frame the task as a Markov Decision Process (MDP), elucidating the intricacies of such interactions;
        \item We construct an \revise{rule-based} policy for the generation of data utilized for task guidance, subsequently employing these data to train reinforcement learning agents with demonstration guidance;
        \item We evaluate and compare the performance of demonstration-guided reinforcement learning agents with their traditional counterparts, observing performance dependence on the number of demonstrations available. \revise{We further perform an ablation study to see dependence on the number of demonstrations.}
    \end{itemize}

\begin{figure}
    \centering
    \includegraphics[width=0.7\linewidth]{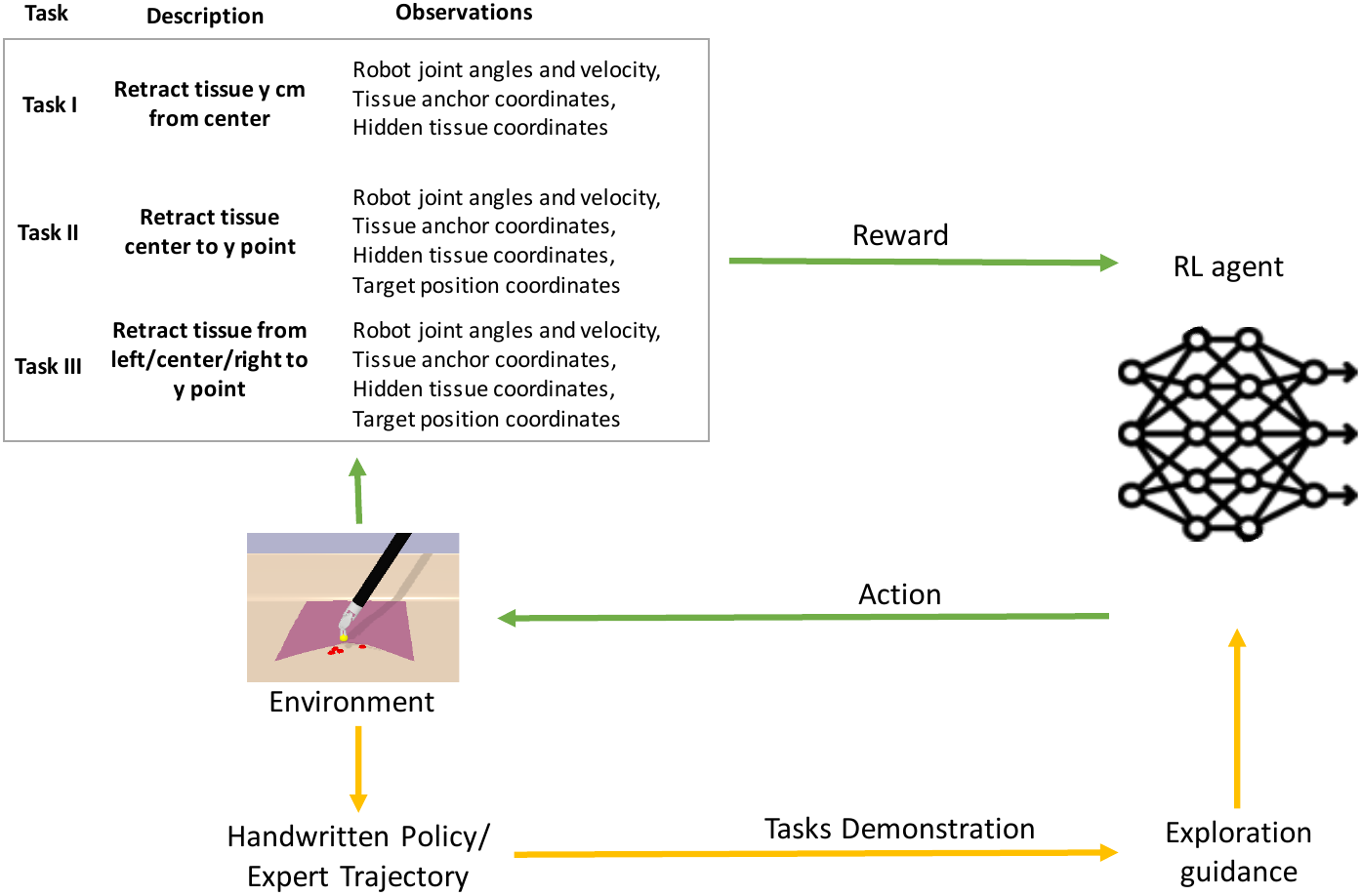}
    \caption{Overview of tasks and pipelines for algorithm training using the environment and \revise{demonstration data} for TissueRetract.}
    \label{fig:overview}
\end{figure}

 \section{Related works} 
The advent of surgical assistive robots has led to a series of pioneering works that have sketched a comprehensive roadmap for surgical autonomy \citep{singh_roadmap_2022}. This transformative concept holds the promise of mitigating surgeon fatigue, reducing operation duration, and facilitating supervised autonomy in telesurgery.
Several robotic systems have emerged, each demonstrating partial autonomy and catering to specific surgical procedures. For example, the CyberKnife system from Accuracy Inc \citep{fuller_high_2022,tree_intensity-modulated_2022} is one such example, utilized predominantly for tumor radiosurgery. Another system, TSolution One, developed by THINK Surgical \citep{liow_think_2017}, is dedicated to rigid bone tissue surgery. ARTAS, a product of Restorative Robots \citep{ARTAS_510k_nodate}, presents the application of robotics in hair restoration. Each of these systems serves as a testament to the tangible benefits of incorporating robotic systems into surgical procedures, paving the way for the progression toward higher surgical autonomy.

Surgical simulation has surged in prominence, finding application in surgical training and the facilitation of task learning. By offering an affordable training modality coupled with the prospect of robust domain randomization, simulation underscores an effective approach to the instruction and enhancement of surgical skills.
The introduction of 3D models, such as the dVRK-VREP simulator \citep{fontanelli_v-rep_2018}, has spurred a substantial exploration of in situ surgical task learning. These investigations cover a diverse array of tasks, including many derived from the Fundamentals of Laparoscopic Surgery (FLS). The advent of the AMBF platform \citep{varier_ambf-rl_2022} has further enriched this field, providing dynamic environments to facilitate interactions between rigid entities and medical robots.
Complementary developments include the Surrol library by Xu et al. \citep{xu_surrol_2021}, specifically designed for rigid body surgical tasks. These include but are not limited to gauze retrieval, needle pickup, and needle retrieval, thereby broadening the range of simulated surgical tasks that can be practiced and mastered.

%SOFT TISSUE SURGICAL TASK LEARNING: and BENEFITS OF SIMULATION:
However, soft tissue manipulation tasks like tissue retraction are a predominant portion of surgical procedures to explore and reach the region of interest (e.g., tumor, gall bladder). 
The task of tissue retraction demands careful gripping and pulling of tissue, to enhance the visibility of obscured regions without inducing tissue damage.
This manipulation of soft tissues is decidedly challenging due to the need for sophisticated tissue tracking and precise planning within the context of dynamically deformable environments. The complexity of the task is further compounded by the requirements for high maneuverability, visual constraints, and the need for repeatability of movements. Such challenges highlight the need for advanced methodologies and tools to effectively navigate these complex surgical procedures.

\revise{For learning soft tissue manipulation, Attanasio et al. use deep neural networks to analyze phantom images along with procedural algorithms and perpendicular retraction gesture, resulting in a 25\% increase in background area after retraction. Murali et al. use learning by observation (LBO), an IL method that uses human demonstrations, followed by a finite state machine (FSM). Saeidi et al. use their Smart Tissue Autonomous Robot (STAR) system for anastomosis in phantoms and in-vivo intestinal tissues. 
To test these algorithms, Murali et al. \citep{murali_learning_2015} use viscoelastic tissue phantoms in conjunction with the Da Vinci Research Kit (dVRK) \citep{kazanzides_open_source_2014}, which facilitates the learning of multilateral tissue cutting. Similarly, Attanasio et al. \citep{attanasio_autonomous_2020} investigated the use of Thiel embalmed cadaverautomating tissue retraction in minimally invasive surgery. Saeidi et al. \citep{saeidi_autonomous_2022} conceived a system for autonomous robotic laparoscopic surgery tailored for intestinal anastomosis, using phantom intestinal tissues and genuine porcine tissue.}

\revise{Soft tissue simulations offer a promising avenue for developing such systems before real-world testing. Tagliabue et al. \citep{tagliabue_soft_2020} introduced UnityFlexML, a Unity-based system, designed to interact with the patient-side manipulator (PSM) through soft body interactions. Soft tissue simulations enable precise measurement of exerted forces, allowing control of training agents with minimal or no tissue damage, a feature that can prove challenging when using tissue phantoms. With advances in physics engines, simulations can also allow for domain randomization to vary tissue color, size, location, tissue
elasticity, and gravity. As such, simulations offer a valuable and scalable tool in advancing the field of surgical robotics.}

\section{Preliminaries}
% RL algorithms, Imitation learning, Demonstration guided RL
In reinforcement learning (RL), agents learn to explore an environment according to the received feedback signals known as rewards.
Reward shaping is a manual task that requires expert knowledge and task-specific fine-tuning, whereas exploration in a sparse reward setting is usually challenging. 
Deep deterministic policy gradient (DDPG)~\citep{ddpg} is an off-policy RL algorithm to mitigate this concern via simultaneously learning the Q-function 
and a policy, using earlier to learn the latter. Q-function $Q_{(s, a)}$ or action value function gives the expected return on taking an arbitrary action at state s.   
Since the Q function is differentiable with respect to action space for environments with continuous action spaces, this allows gradient-based learning of policy.
In DDPG, for any D set of transitions $(s_t,a,r,s_{t+1},d)$ collected, the training objective of the agent is to learn the policy $\mu_{\theta}(s)$ whose actions maximize $Q_\phi(s,a), a\sim \mu_\theta(s)$ and the Q function as follows:

\begin{equation}
    \max_{\theta} \underset{s \sim D,\ a\sim \mu_\theta(s)}{\mathbb{E}} [Q_\phi (s, a)].
\end{equation}

Imitation learning (IL), also known as learning from demonstration, is an approach to learning behavior from given \revise{demonstration data}. Instead of manually designing reward functions, the reward function is learned from demonstrations. Imitation learning algorithms can learn via supervised learning to mimic demonstrations (behavior cloning) or directly decode rewards from demonstrations to learn policy (inverse reinforcement learning). \citep{reddy_sqil_2020} IL struggles to learn robust policies for changing environment parameters, and joint optimization of reward and policy makes it difficult to train. Soft Q imitation learning (SQIL) is a model-free off-policy algorithm that can solve for both discrete and continuous action spaces \citep{reddy_sqil_2020}. 

Demonstration-guided reinforcement learning (DG-RL) combines imitation learning with reinforcement learning. These methods can provide one or more of the following: better initialization of the agent using BC, augment demonstrations into replay buffer to improve the signal or augment environment reward with demonstration guide reward. \citep{ddpgbc, col,dex}.
\revise{DDGPBC \citep{ddpgbc} algorithm was developed to address the exploration in environments with sparse rewards, by using demonstrations to successfully learn long-horizon, multi-step robotics tasks with continuous control.} Cycle-of-learning (CoL) \citep{col} is an actor-critic-based method that transitions from behavior cloning to reinforcement learning. It combines pre-training and joint loss functions to learn both value and policy functions from demonstrations. Demonstration-guided exploration (DEX) \citep{dex} is a DG-RL method that improves potential overestimation over previous actor-critic methods and augments environment rewards with a behavior gap between agent and expert policy. \revise{Several new algorithms have been proposed for each of these categories. For proof of concept, we limit our work to these algorithms only.}

\section{Methods}
In this section, we describe the TissueRetract environment and formulate the tissue retraction task as a Markov decision process. We discuss success criteria and the implementation of reinforcement learning algorithms used.
\revise{Figure \ref{fig:overview} shows a graphic overview of the training pipeline and components of the environment tasks.}

% Describe the environment and scenarios
\subsection{Environment}
We propose the TissueRetract environment as a new benchmark for surgical soft tissue manipulation tasks. In this work, we simulate a patient-side manipulator (PSM) robot as a rigid body with 6 degrees of freedom (DOF) and the tissue as a soft body, with hidden tissue lying underneath. We define anchor points to guide the robotic arm to hold the soft tissue. To stabilize the soft tissue, it is fixed at all corners. To succeed at a task, the agent needs to learn to hold the tissue from anchor points and pull it until success criteria are met. Our simulation environment is ROS-compatible and follows the OpenAI gym structure. We use 3D models from dVRK-VREP simulator \citep{fontanelli_v-rep_2018}, and \revise{PyBullet physics engine} \citep{pybullet} for the simulation of rigid and soft body physics.

We introduce three \revise{tasks with varying levels of difficulty. Two important variables considered for tissue retraction are tissue retraction distance and the anchor site for tissue holding. We design these three tasks with a stepwise increase in variability on one of the two variables' levels.} The \revise{task I} requires vertical manipulation of soft tissue using an anchor point until a distance less than the threshold is reached. The \revise{task II} requires soft tissue manipulation of anchor points given target instructions until a distance less than the distance threshold is reached. The \revise{task III} involves manipulation based on one of the multiple anchor points until the distance threshold is reached. \revise{Robot location is randomly sampled from an arbitrary uniform distribution. Tissue location coordinates \((X, Y)\), are sampled independently from a uniform distribution as $X \sim \text{U}(x_{\text{min}}, x_{\text{max}}), \quad Y \sim \text{U}(y_{\text{min}}, y_{\text{max}})$, where $\text{U}$ is a uniform distribution and \([x_{\text{min}}, x_{\text{max}}, y_{\text{min}}, y_{\text{max}}]\) are x and y boundaries of robot's workspace projected on a 2d surface.}

\begin{figure}
    \centering
    \includegraphics[width=0.9\linewidth]{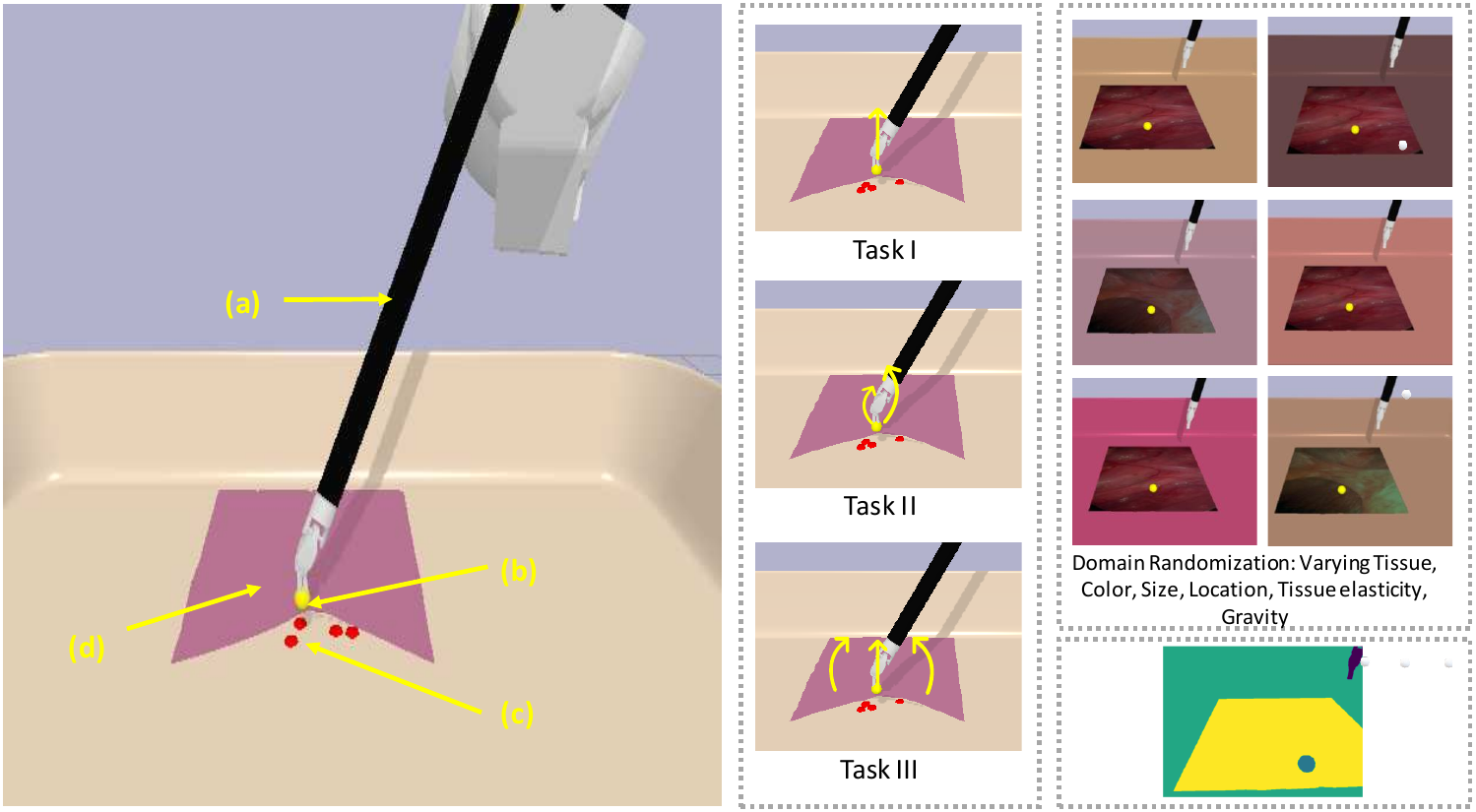}
    \caption{Overview of tasks in TissueRetract environment Left: Components of simulation. a = DaVinci PSM arm, b = Target point, c= Hidden Tissue, d = Soft tissue. Middle Top: Task I - Pull tissue from center, Middle Center: Task II - Pull tissue to target point x from center, Middle Bottom: Task III - Pull tissue to target point x from center/left/right. Right Top: Physics simulation engines allow for domain randomization in tissue color, texture, size, location, and tissue elasticity. Right Bottom: Segmentation mask of components.}
    \label{fig:envirnment}
\end{figure}

\revise{Figure \ref{fig:envirnment} shows a graphic about components of the environment and a pictorial representation of the three tasks.}

% Scientific problem formulation
\subsection{Problem Formulation}
We frame our task as a Markov Decision Process (MDP) where an agent learns by interacting with the environment. An MDP is defined as a tuple $\langle \mathcal{S}, \mathcal{A}, R, P, \gamma \rangle$, 
where $\mathcal{S}$ indicates the state space, 
$\mathcal{A}$ indicate the action space of the agent, 
$R:\mathcal{S}\times \mathcal{A}\to\mathbb{R}$ is the reward function given the states and actions, 
$P$ indicates the probability function of state transitions $\mathcal{S}\times\mathcal{A}\to\mathcal{S}$, 
and $\gamma$ is the discount factor to penalize long action sequences. 
At each time step $t$, the agent will take action $a_t\sim\mu_\theta(s_t)$ based on the learned policy $\mu_\theta$ and the current agent state $s_t$.
With the current state $s_t$ and the action $a_t$, the agent will transit to the next state $s_{t+1}$ according to the transition probability function $P(s_{t+1}|s_t,a_t)$ and generate the corresponding reward $r_t=R(s_t,a_t)$.
The Agent experience $E_t$=($s_t, a_t, r_t, s_{t+1}$) from beginning till episode completion, 
is stored in replay buffer $D_A$. The episode is completed either on successful completion of the task or reaching the time horizon. The \revise{demonstration data is} stored as experience in another buffer $D_E$. Depending on the agent, it can learn from $D_A$, $D_E$, or both. The agent's goal is to learn the policy $\mu: S \rightarrow A$ such that it maximizes the expected reward over a time horizon.

Since reward engineering requires hand tailoring and is non-scalable, we use sparse reward, where the agent only receives binary rewards of whether the task is completed or not.
We define distance-based success criteria that require tissue to be lifted within the error tolerance of the desired height. We use the success rate as a metric to compare algorithms. The success rate is defined as the ratio of the successful completion of the task to the number of attempts.

\subsection{Algorithms} 
We explore traditional reinforcement learning algorithms (DDPG\citep{ddpg}), Imitation learning algorithms (SQIL\citep{reddy_sqil_2020}), and demonstration-guided RL algorithms (DDPGBC\citep{ddpgbc}, CoL\citep{col}, and DEX\citep{dex}). 
\revise{Each algorithm is trained for a total of 50,000 episodes. For the IL and DG-RL methods, we restrict the demonstration count to 100. We conducted our experiments on a single Intel i7-10750H CPU with a NVIDIA GeForce GTX 1650 Ti. In addition, we performed an ablation study at 25, 50, and 100 demonstrations to compare the effect of demonstration numbers on the IL and DG-RL algorithms.}
For actor and critic networks in DDPG, we use three fully connected layers, each of 128 dimensions with RELU activation. We use hindsight experience replay with a future sampling strategy, with a buffer size of 50k observations, a discount factor of 0.99, and a learning rate of 0.001, along with adam optimizer for model optimization. The episode horizon is set to 50 time steps.  For SQIL, we initialize demonstration data with reward 0, and as agents interact with the environment, add new experiences with the reward of -1 to the replay buffer. For DG-RL algorithms, we use DDPG as the base and add demonstration data to the replay buffer during each minibatch. For DDPGBC, we use behavior cloning loss along with Q-filter. In CoL, we augment behavior cloning loss with actor Q loss from DDPG. Similar to the original work, we implement DEX with DDPG with four fully connected layers of 256 dimensions each. At the start of each simulation, the PSM arm, soft tissue, and hidden tissue locations are randomly chosen from a uniform distribution. 

\revise{During the evaluation, we run each algorithm for 50 episodes, each with a horizon of 50 timesteps. Similar to Huang et al. \citep{dex}, evaluation is repeated 3 times with varying random seeds of 1,2 and 3. The final average success rate, along with a 95\% confidence interval is reported. We also compare the effect of demonstration count on the success rate percentage. For this, we calculate a 95\% confidence interval for success rate percentage over 50 episodes with varying seed values for IL and DG-RL algorithms, when trained on 25, 50, and 100 demonstrations. Finally, we also perform a manual visual inspection of success and failure cases across the model training journey and look for patterns in behavior emergence. For this, we sample agent evaluations after every 10,000 training episodes, manually inspect the cause of failure, and classify the causes. We discuss more about this in section 5.2.}

\subsection{Demonstration Data Generation} 
For demonstration guidance, we generate rule-based \revise{demonstration data}. We define the anchor point, a PSM, and soft tissue with respect to the frame of origin. We further break down the task into a sequence of 4 position checkpoints: Approaching the tissue, holding the tissue, tissue retraction until the threshold is met, and finally holding the tissue. Using Inverse kinematics, we derive the required action to move the agent to the required checkpoint. The agent experience is stored as ($s_t, a_t, s_{t+1}$) per time step. This can be replaced by real data collected from surgical procedures.

\begin{table*}[!t]%
\centering %
\caption{Comparison of success rate \revise{percentage, over 50 evaluation episodes} using DDPG, SQIL, DDPGBC, CoL, and DEX for tasks I, II, and III. We fixed the demonstration count at 100, \revise{and trained all algorithms with random seeds of 1,2 and 3 respectively to calculate 95\% confidence interval for success rate percentage. DDPG achieved comparable or higher success rates than DG-RL algorithms (average success rates of 85, 84, and 66 on tasks I, II, and III). SQIL baseline fails on all three tasks.}}

\begin{tabular}{c|c|c|ccc}\toprule
\textbf{}        & \textbf{\textbf{RL}} & \textbf{\textbf{IL}} & \multicolumn{3}{c}{\textbf{DG-RL}} 
\\
\cmidrule{2-6} \textbf{Task}  & DDPG                 & SQIL                  & CoL        & DDPGBC    & DEX      \\ \hline
Task I               & 83$\pm$0.7           & 4$\pm$7               & 94$\pm$6   & 95$\pm$3  & 66$\pm$32   \\
Task II               & 85$\pm$11            & 1$\pm$1               & 86$\pm$9   & 86$\pm$5  & 82$\pm$12  \\
Task III               & 80$\pm$6             & 6$\pm$4               & 76$\pm$4   & 65$\pm$5  & 58$\pm$8   \\
\hline
\end{tabular}
\label{tab:results}
\end{table*}

\section{Results and Discussion}
% In this section, we discuss the performance of different algorithms in all tasks, learning behavior, and dependence on demonstration data.

\begin{table*}[!t]%
\centering %
\caption{Comparison of success rate \revise{percentage, over 50 evaluation episodes using 25, 50, and 100 demonstrations on task III. We trained all algorithms with random seeds of 1,2 and 3 respectively to calculate a 95\% confidence interval for the success rate percentage. Demonstration-guided RL algorithms (DDPGBC, CoL, and DEX) benefit from an increase in the number of demonstrations available, with improvement in either the average or the bounds of the success rate. SQIL benefits slightly, without any significant gains in performance.}}
\begin{tabular}{cccc} \toprule
       & \multicolumn{3}{c}{\textbf{Number of demonstrations}}                                   \\ \cmidrule{2-4}
\textbf{Method} & \multicolumn{1}{c}{\textbf{25 demos}} & \multicolumn{1}{c}{\textbf{50 demos}} & \textbf{100 demos} \\ \hline
SQIL   & \multicolumn{1}{c}{5$\pm$1}       & \multicolumn{1}{c}{5$\pm$1}       &   6$\pm$4       \\
CoL    & \multicolumn{1}{c}{50$\pm$7}       & \multicolumn{1}{c}{60$\pm$18}       &   76$\pm$4  \\ 
DDPGBC & \multicolumn{1}{c}{62$\pm$7}       & \multicolumn{1}{c}{62$\pm$7}       &   65$\pm$5      \\ 
DEX    & \multicolumn{1}{c}{34$\pm$10}       & \multicolumn{1}{c}{58$\pm$16}       &   58$\pm$8      \\ 
\hline
\end{tabular}
\label{tab:ablation_results}
\end{table*}

\begin{figure}
\centering
\resizebox*{8cm}{!}{\includegraphics{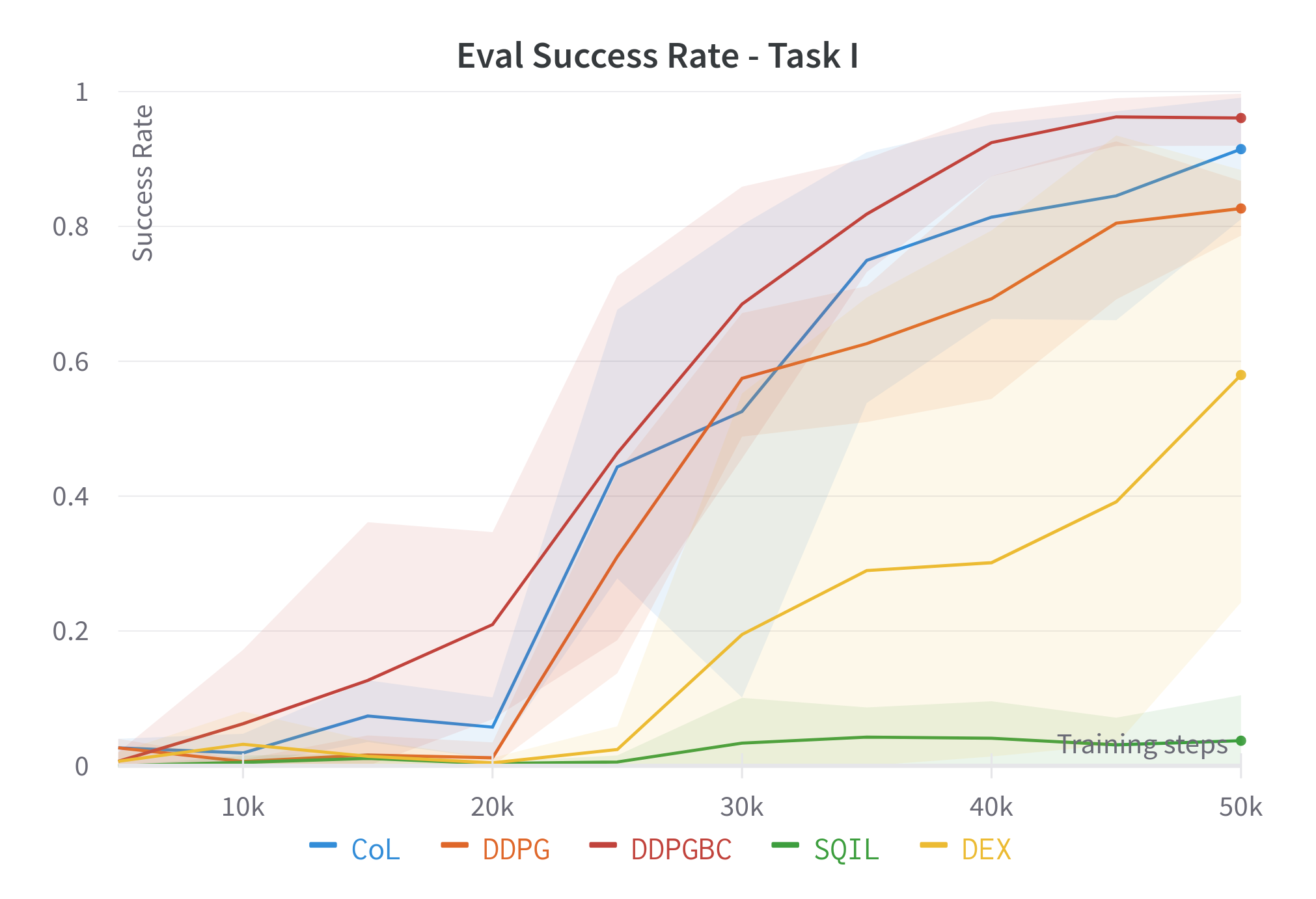}}\hspace{3pt}
\resizebox*{8cm}{!}{\includegraphics{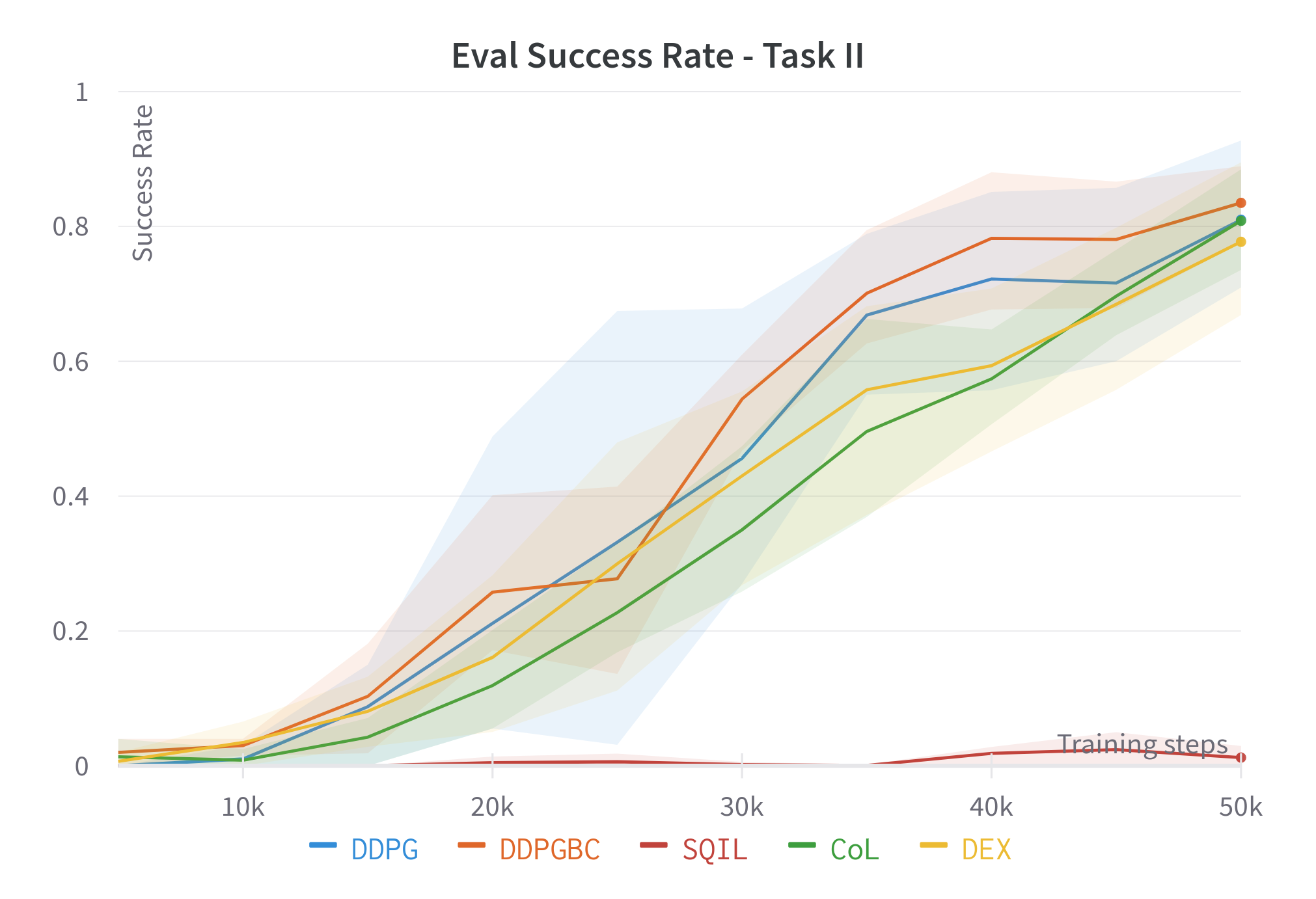}}\hspace{3pt}
\resizebox*{8cm}{!}{\includegraphics{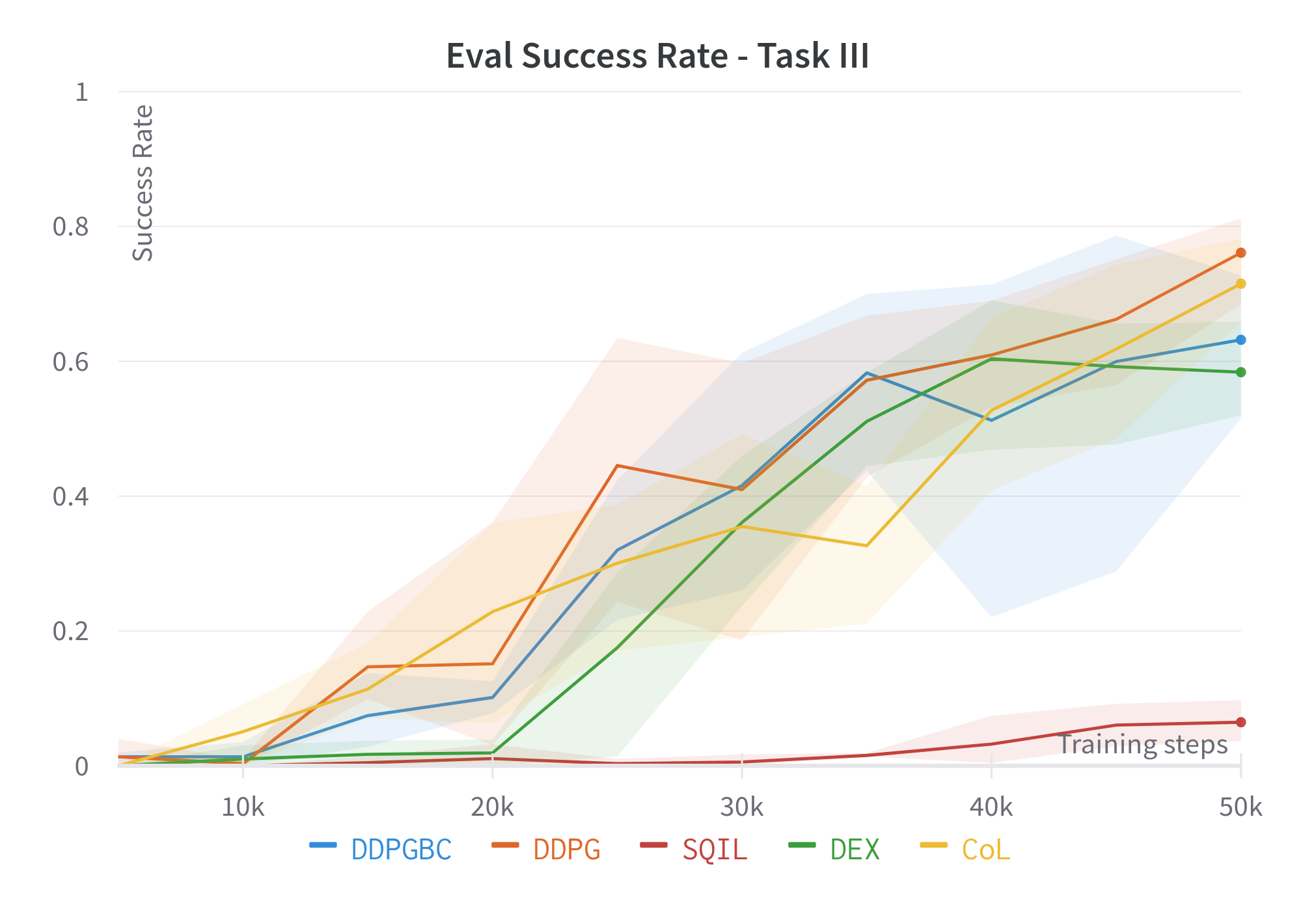}}
\resizebox*{8cm}{!}{\includegraphics{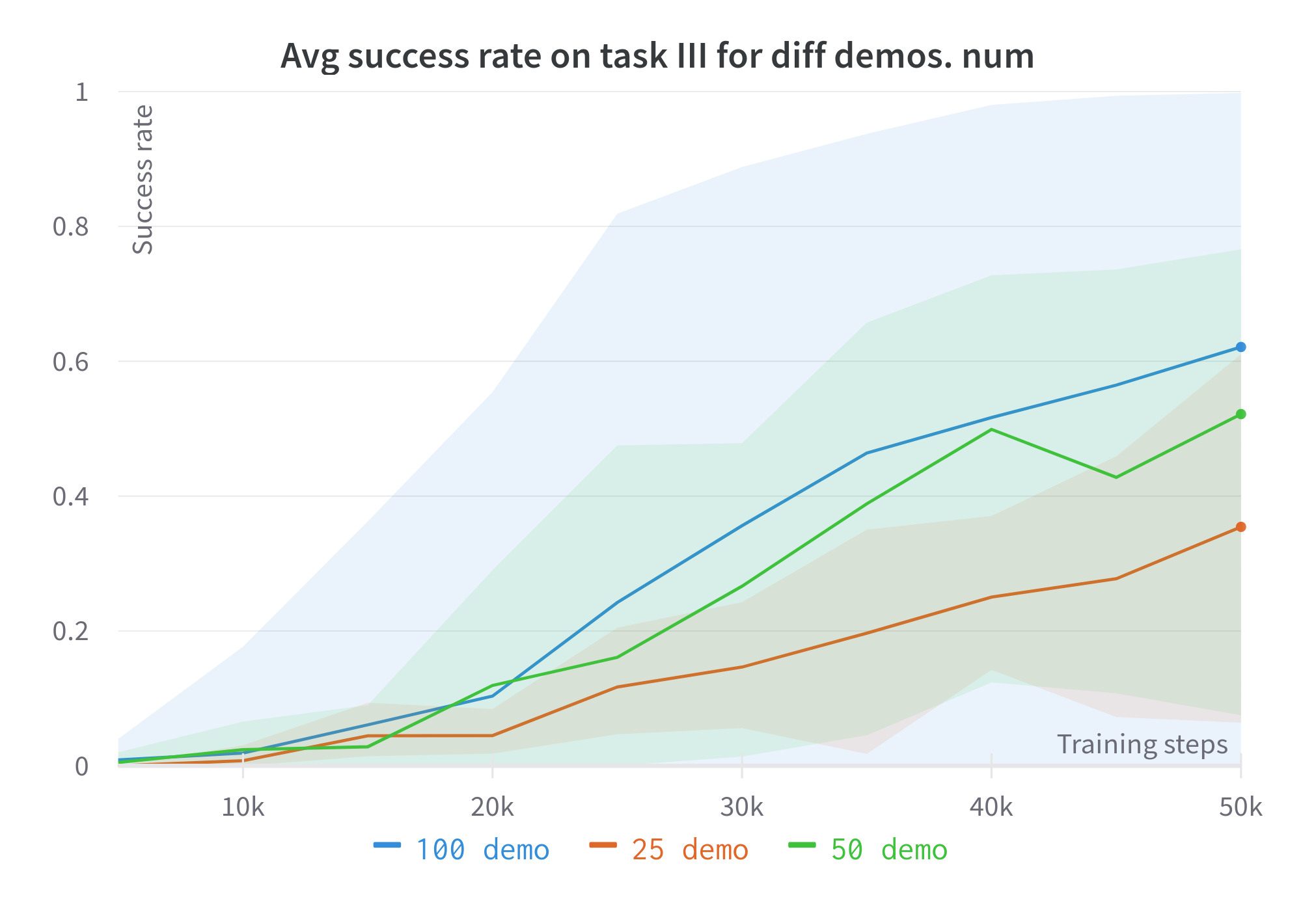}}\hspace{3pt}
\caption{Performance curves for three tasks using DDPG, SQIL, DDPGBC, CoL, and DEX. Top Left: Evaluation success rate on \revise{task I grouped by algorithm used, Top Right: Evaluation success rate on task II grouped by algorithm used, Bottom Left: Evaluation success rate on task III grouped by algorithm used, Bottom Right: Average success rate of all 5 algorithms,} grouped by number of demonstrations.} 
 \label{fig:results}
\end{figure}

\subsection{Agent Performance}
\revise{Figure \ref{fig:results} shows the changes in the success rate as the model training progresses. In our initial results, we achieved an average success rate percentage of 69, 68, and 57 on tasks I, II, and III respectively. SQIL remains an outlier, with a near-zero success rate across all tasks. This is possibly due to the large exploration space of the task, which might not be possible directly by observing observations. On excluding SQIL from calculations, we have achieved an average success rate percentage of 85, 85, and 70 on tasks I, II, and III respectively. DDPG achieved comparable or higher success rates than DG-RL algorithms (average success rate of 85, 84, and 66 on tasks I, II, and III). This shows that pure explorations in RL algorithms can compete or surface demonstration-guided exploration for DG-RL algorithms.} 

\begin{figure*}
\centerline{\includegraphics[width=\linewidth]{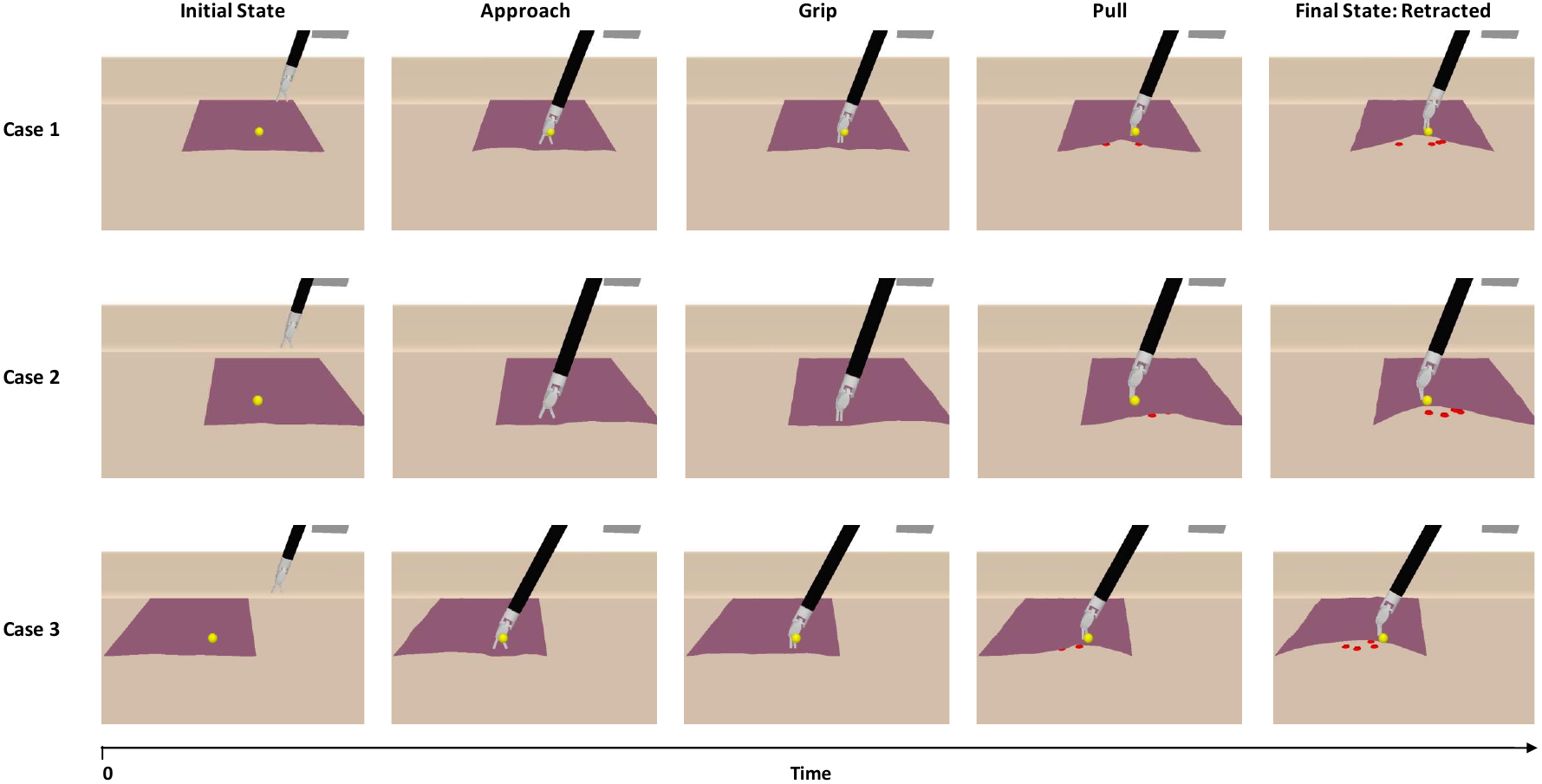}}
\caption{Case studies of success from \revise{task III}, with varying anchor points. Case 1: Middle anchor point; Case 2: Left anchor point; Case 3: Right anchor point. After successful training, the agent starts by approaching the tissue, grasping the anchor point, and pulling it till retraction is complete.}
\label{fig:normalcases}
\end{figure*}

\revise{Figure \ref{fig:normalcases} shows a snapshot of post-training three successful episodes with varying anchor sites, and phases for task completion namely: initialization, approaching, grip, pull, and maintaining retraction. Initialization is the first phase, where all environment variables are set up. In the second phase, the agent approaches the anchor sire. In third phase involves robot aligning followed by gripping of tissue. In the fourth phase, the agent pulls the tissue followed by maintaining the retraction in the final state. The performance of all five algorithms in all three tasks is shown in Table \ref{tab:results}.  
}

\subsection{Initial Behavior and Failure cases}
We investigate the evolution of agent behavior by sampling episodes as training progresses. The agent initially learns to explore the region and approach the tissue, grips the tissue from the anchor points, and finally learns to retract. \revise{These patterns of behavior emergence can explain the causes of failure cases after the training. We observe three common failure scenarios: (a) improper grasp, where the anchor point moves with the soft tissue, making grasping difficult. (b) tissue distortion: agents mishandle robots resulting in tissue distortion and possible damage; and (c) loss of tissue grip, due to recoil on stretching soft tissue. This can be also due to sudden retraction by the agent, leading to loss of tissue grip. These behaviors can be possibly improved by training the agent for longer time steps, increasing demonstration counts, or introducing a specific negative reward for tissue damage.}

\begin{figure*}
\centerline{\includegraphics[width=\linewidth]{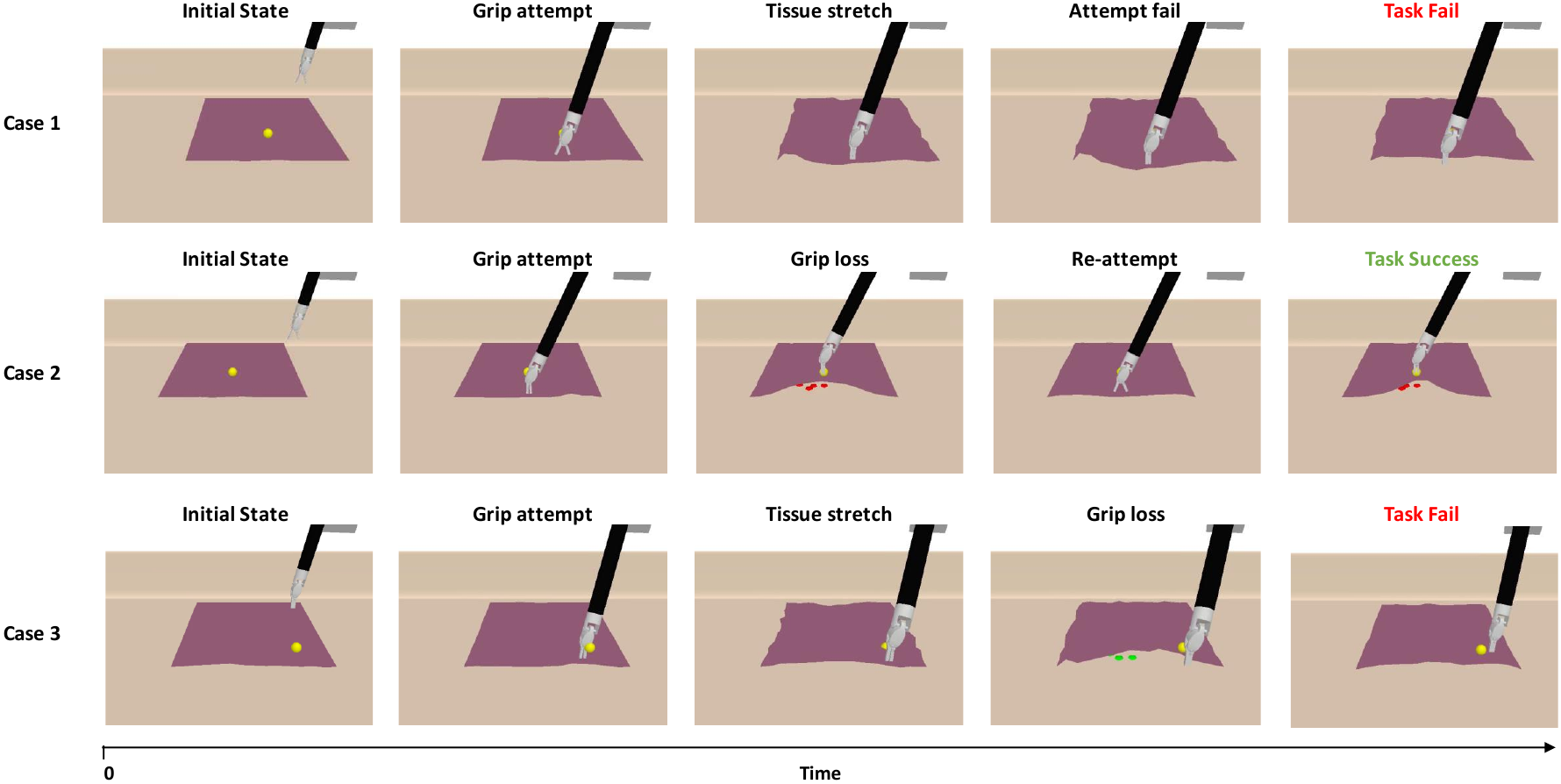}}
\caption{Case studies with common abnormal variants from \revise{task III}. The agent learns to reattempt cases of grip loss. Case 1: Task fails as tissue slips away due to tissue elasticity; Case 2: Grip loss, succeed on reattempt; Case 3:Grip loss, success of reattempt.}
\label{fig:abnormalcases}
\end{figure*}

\revise{Figure \ref{fig:abnormalcases} shows a snapshot of failed episodes after model training. Case 1 of Figure \ref{fig:abnormalcases} shows failure in tissue gripping as tissue slips away due to the elasticity of tissue. Case 2 shows the loss of tissue grip while retraction, followed by success in reattempting to grip the tissue.  Case 3 shows similar scenarios where the agent fails on a similar reattempt. More in-depth analysis is required to understand the correlation between causes of failure and its emergence in behavior sequence during learning.}

\subsection{Effect of Number of Demonstrations}
\revise{Table \ref{tab:ablation_results} shows the ablation results for the number of demonstrations on DG-RL and IL algorithms. Since DGGP doesn't use the demonstration data, we will only discuss the rest of the algorithms. In general, increasing the demonstration count has a positive effect on agent performance, resulting in an increased success rate percentage or narrowing of performance bounds over several runs.  Demonstration-guided RL algorithms (DDPGBC, CoL, and DEX) benefit from an increase in the number of demonstrations available, with the improvement of average success rates by 10 and 6 percent on the addition of 25 and 50 additional demonstrations. SQIL benefits slightly, without any significant gains in performance. The bottom right subfigure of Figure \ref{fig:results} shows the algorithm's success rate average. It is interesting to note that despite changes in magnitude, the changes in peak are fairly similar across a number of demonstrations provided. This can be possibly hint that the number of demonstrations affects the final success rate percentage, and doesn't impact the rate of learning new tasks.}  

\subsection{Limitations and Future Works}
\revise{There are several limitations of the existing work. For a more comprehensive analysis, it would be further beneficial to conduct longer training episodes. In our experiments, we find that pure explorations in RL algorithms can compete or surface demonstration-guided exploration in DG-RL algorithms. Further experimentation is required to access this in depth. Another dimension of analysis is required to understand the correlation between causes of failure and its emergence in behavior sequence during learning. Although the present study intentionally eschewed explicit reward engineering, the induction of certain preferred traits in agent behavior may require customized reward engineering. For example, training an RL agent to expedite soft tissue manipulation, follow the shortest path, or eliminate tissue damage can require the integration of additional information within both the observation and reward structures provided to the agent. Such a direction presents an intriguing avenue for future exploration.
Furthermore, additional testing is required to validate the robustness of the transferability from simulation to reality. The scope of this work could also be expanded to incorporate vision-based control, surgical mesh fixation, or a multitask learning framework. These potential extensions of our work could further enhance the capabilities of surgical soft tissue manipulation, enriching its applicability to surgical robotics.}

\section{Conclusion} 
In conclusion, this work presents a proof-of-concept study of soft tissue simulation with rigid body interactions, demonstrating a meaningful advance in the simulation of surgical tasks. By employing the PyBullet physics engine, we replicate the kinematics of the patient-side manipulator to simulate soft and rigid body interactions. Additionally, through the use of demonstration guidance, we train reinforcement learning agents to master the task. After training, the agents \revise{were able to execute all 3 tasks with high success rates.}
Our research provides an innovative approach for autonomous soft surgical tissue retraction. In addition, it introduces a comprehensive framework for the in silico learning of surgical tasks with soft tissue manipulation. This modality of in silico training, followed by sim-to-real transfer, has the potential to significantly broaden access to soft tissue manipulation research. It also emerges as a practical and expedient approach to rapid prototyping of automated surgical procedures. 

\section*{Code Availability} 
The code is available at \href{https://github.com/amritpal-001/tissue_retract}{https://github.com/amritpal-001/tissue\_retract}.


\begin{thebibliography}{10}
\providecommand \doibase [0]{http://dx.doi.org/}%

\bibitem{Hirt1974}
Hirt CW, Amsden AA, Cook JL. An arbitrary {L}agrangian-{E}ulerian computing
  method for all flow speeds. {\it J {C}omput {P}hys.}
  1974\string;14(3)\string:227--253.

\bibitem{Liska2010}
Liska R, Shashkov M, Vachal P, et al. Optimization-based synchronized
  flux-corrected conservative interpolation (remapping) of mass and momentum
  for arbitrary {L}agrangian-{E}ulerian methods. {\it J {C}omput {P}hys.}
  2010\string;229(5)\string:1467--1497.

\bibitem{Taylor1937}
Taylor GI, Green AE. Mechanism of the production of small eddies from large
  ones. {\it P {R}oy {S}oc {L}ond {A} {M}at.}
  1937\string;158(895)\string:499--521.
\newblock \url{https://doi.org/10.1098/rspa.1937.0036},
  \url{http://rspa.royalsocietypublishing.org/content/158/895/499}.

\bibitem{Knupp1999}
Knupp PM. Winslow smoothing on two-dimensional unstructured meshes. {\it Eng
  {C}omput.} 1999\string;15\string:263--268.

\bibitem{Kamm2000}
Kamm J. Evaluation of the {S}edov-von {N}eumann-{T}aylor blast wave solution.
  Tech. Rep. Technical {R}eport LA-UR-00-6055, Los {A}lamos {N}ational
  {L}aboratory; The address:   2000.

\bibitem{Kucharik2003}
Kucharik M, Shashkov M, Wendroff B. An efficient linearity-and-bound-preserving
  remapping method. {\it J {C}omput {P}hys.}
  2003\string;188(2)\string:462--471.

\bibitem{Blanchard2015}
Blanchard G, Loubere R. High-Order {C}onservative {R}emapping with a posteriori
  {MOOD} stabilization on polygonal meshes. Details on how published;  2015.
\newblock Accessed January 13, 2016.
  \url{https://hal.archives-ouvertes.fr/hal-01207156}, the {HAL} {O}pen
  {A}rchive, hal-01207156.

\bibitem{Burton2013}
Burton DE, Kenamond MA, Morgan NR, Carney TC, Shashkov MJ, Author AB. An
  intersection based {ALE} scheme {(xALE)} for cell centered hydrodynamics
  {(CCH)}. In: Talk at {M}ultimat 2013, {I}nternational {C}onference on
  {N}umerical {M}ethods for {M}ulti-{M}aterial {F}luid {F}lows. The
  Organization.  September 2--6, 2013; San {F}rancisco.
\newblock LA-UR-13-26756.2.

\bibitem{Berndt2011}
Berndt M, Breil J, Galera S, Kucharik M, Maire PH, Shashkov M. Two-step hybrid
  conservative remapping for multimaterial arbitrary {L}agrangian-{E}ulerian
  methods. {\it J {C}omput {P}hys.} 2011\string;230(17)\string:6664--6687.

\bibitem{Kucharik2012}
Kucharik M, Shashkov M. One-step hybrid remapping algorithm for multi-material
  arbitrary {L}agrangian-{E}ulerian methods. {\it J {C}omput {P}hys.}
  2012\string;231(7)\string:2851--2864.

\bibitem{Breil2015}
Breil J, Alcin H, Maire PH. A swept intersection-based remapping method for
  axisymmetric {ReALE} computation. {\it Int {J} {N}umer {M}eth {F}l.}
  2015\string;77(11)\string:694--706.
\newblock Fld.3996.

\bibitem{Barth1997}
Barth TJ. Numerical methods for gasdynamic systems on unstructured meshes. In:
  Kroner D, Rohde C, Ohlberger M. \kern-2pt, eds. {\it An {I}ntroduction to
  {R}ecent {D}evelopments in {T}heory and {N}umerics for {C}onservation {L}aws,
  {P}roceedings of the {I}nternational {S}chool on {T}heory and {N}umerics for
  {C}onservation {L}aws}, 2~ed., Lecture {N}otes in {C}omputational {S}cience
  and {E}ngineering. Springer,  1997.

\bibitem{Lauritzen2011}
Lauritzen P, Erath C, Mittal R. On simplifying `incremental remap'-based
  transport schemes. {\it J {C}omput {P}hys.}
  2011\string;230(22)\string:7957--7963.

\bibitem{Klima2017}
Klima M, Kucharik M, Shashkov M. Local error analysis and comparison of the
  swept- and intersection-based remapping methods. {\it Commun {C}omput
  {P}hys.} 2017\string;21(2)\string:526--558.

\bibitem{Dukowicz2000}
Dukowicz JK, Baumgardner JR. Incremental remapping as a transport/advection
  algorithm. {\it J {C}omput {P}hys.} 2000\string;160(1)\string:318--335.

\bibitem{Kucharik2011}
Kucharik M, Shashkov M. Flux-based approach for conservative remap of
  multi-material quantities in {2D} arbitrary {L}agrangian-{E}ulerian
  simulations. In:  Fo\v{r}t J, F{\"{u}}rst J, Halama J, Herbin R, Hubert F.
  \kern-2pt, eds. {\it Finite {V}olumes for {C}omplex {A}pplications {VI}
  {P}roblems \& {P}erspectives}, 1~ed., Springer {P}roceedings in
  {M}athematics. Springer,  2011\string:623--631.

\bibitem{Kucharik2014}
Kucharik M, Shashkov M. Conservative multi-material remap for staggered
  multi-material arbitrary {L}agrangian-{E}ulerian methods. {\it J {C}omput
  {P}hys.} 2014\string;258\string:268--304.

\bibitem{Loubere2005}
Loubere R, Shashkov M. A subcell remapping method on staggered polygonal grids
  for arbitrary-{L}agrangian-{E}ulerian methods. {\it J {C}omput {P}hys.}
  2005\string;209(1)\string:105--138.

\bibitem{Caramana1998}
Caramana EJ, Shashkov MJ. Elimination of artificial grid distortion and
  hourglass-type motions by means of {L}agrangian subzonal masses and
  pressures. {\it J {C}omput {P}hys.} 1998\string;142(2)\string:521--561.

\bibitem{Hoch2009}
Hoch P. An arbitrary {L}agrangian-{E}ulerian strategy to solve compressible
  fluid flows. Tech. Rep. Technical {R}eport, CEA; The address:   2009.
\newblock Accessed January 13, 2016. HAL: hal-00366858.
  https://hal.archives-ouvertes.fr/docs/00/36/68/58/PDF/ale2d.pdf.

\bibitem{Shashkov1996}
Shashkov M. {\it Conservative {F}inite-{D}ifference {M}ethods on {G}eneral
  {G}rids}.
\newblock CRC {P}ress, 1996.

\bibitem{Benson1992}
Benson DJ. Computational methods in {L}agrangian and {E}ulerian hydrocodes.
  {\it Comput {M}ethod {A}ppl {M}.} 1992\string;99(2--3)\string:235--394.

\bibitem{Margolin2003}
Margolin LG, Shashkov M. Second-order sign-preserving conservative
  interpolation (remapping) on general grids. {\it J {C}omput {P}hys.}
  2003\string;184(1)\string:266--298.

\bibitem{Kenamond2013}
Kenamond MA, Burton DE. Exact intersection remapping of multi-material
  domain-decomposed polygonal meshes. In: Talk at {M}ultimat 2013,
  {I}nternational {C}onference on {N}umerical {M}ethods for {M}ulti-{M}aterial
  {F}luid {F}lows. The Organization.  September 2--6, 2013; San {F}rancisco.
\newblock LA-UR-13-26794.

\bibitem{Dukowicz1984}
Dukowicz J. Conservative rezoning (remapping) for general quadrilateral meshes.
  {\it J {C}omput {P}hys.} 1984\string;54(3)\string:411--424.

\bibitem{Margolin2002}
Margolin LG, Shashkov M. Second-order sign-preserving remapping on general
  grids. Tech. Rep. Technical Report LA-UR-02-525, Los {A}lamos {N}ational
  {L}aboratory; The address:   2002.

\bibitem{Mavriplis2003}
Mavriplis DJ. Revisiting the least-squares procedure for gradient
  reconstruction on unstructured meshes. In: AIAA 2003-3986. 16th {AIAA}
  {C}omputational {F}luid {D}ynamics {C}onference. The organization.  June
  23--26, 2003; Orlando, {F}lorida.

\bibitem{Scovazzi2008}
Scovazzi G, Love E, Shashkov M. Multi-scale {L}agrangian shock hydrodynamics on
  {Q1/P0} finite elements: {T}heoretical framework and two-dimensional
  computations. {\it Comput {M}ethod {A}ppl {M}.}
  2008\string;197(9--12)\string:1056--1079.

\end{thebibliography}


\begin{thebibliography}{10}
% \providecommand \doibase [0]{http://dx.doi.org/}%
\bibitem{mylesGlobal2010} Myles PS, Haller G. Global distribution of access to surgical services. The Lancet. 2010;376(9746):1027–1028. Publisher: Elsevier doi: 10.1016/S0140-6736(10)60520-X 

\bibitem{debasEmergence2015}  Debas HT. The Emergence and Future of Global Surgery in the United States. JAMA Surgery. 2015;150(9):833–834. doi: 10.1001/jama-surg.2015.0898

\bibitem{sturm_effects_2011} Sturm L, Dawson D, Vaughan R, et al. Effects of fatigue on surgeon performance and surgical outcomes: a systematic review. ANZ Journal of Surgery.2011;81(7-8):502–509. \_eprint: https://onlinelibrary.wiley.com/doi/pdf/10.1111/j.1445-2197.2010.05642.xdoi: 10.1111/j.1445-2197.2010.05642.x
\bibitem{todorov_mujoco_2012} Todorov E, Erez T, Tassa Y. MuJoCo: A physics engine for model-based control. In: 2012 IEEE/RSJ International Conference on Intelligent Robots and Systems. ; 2012:5026-5033. doi:10.1109/IROS.2012.6386109
\bibitem{pybullet} Coumans E, Bai Y. PyBullet, a Python module for physics simulation for games, robotics and machine learning. http://pybullet.org; 2016–2022.
\bibitem{ARTAS_510k_nodate} ARTAS. 510(k) Premarket Notification. Accessed July 14, 2023. https://www.accessdata.fda.gov/scripts/cdrh/cfdocs/cfpmn/pmn.cfm?ID=K173358
\bibitem{tagliabue_soft_2020} Tagliabue E, Pore A, Dall’Alba D, Magnabosco E, Piccinelli M, Fiorini P. Soft Tissue Simulation Environment to Learn Manipulation Tasks in Autonomous Robotic Surgery. In: 2020 IEEE/RSJ International Conference on Intelligent Robots and Systems (IROS). ; 2020:3261-3266. doi:10.1109/IROS45743.2020.9341710
\bibitem{xu_surrol_2021} Xu J, Li B, Lu B, Liu YH, Dou Q, Heng PA. SurRoL: An Open-source Reinforcement Learning Centered and dVRK Compatible Platform for Surgical Robot Learning. Published online August 30, 2021. doi:10.48550/arXiv.2108.13035
\bibitem{nagy_performance_2022}  Performance and Capability Assessment in Surgical Subtask Automation. Accessed July 2, 2023. https://www.mdpi.com/1424-8220/22/7/2501
\bibitem{singh_roadmap_2022} Singh A. Roadmap to Autonomous Surgery -- A Framework to Surgical Autonomy. Published online May 26, 2022. doi:10.48550/arXiv.2206.10516
\bibitem{cabrelli_stable_2016} Stable phantom materials for ultrasound and optical imaging - IOPscience. Accessed August 23, 2023. https://iopscience.iop.org/article/10.1088/1361-6560/62/2/432
\bibitem{dex}  Huang T, Chen K, Li B, Liu YH, Dou Q. Demonstration-Guided Reinforcement Learning with Efficient Exploration for Task Automation of Surgical Robot. 2023. arXiv:2302.09772 [cs]


\bibitem{attanasio_autonomous_2020} Attanasio A, Scaglioni B, Leonetti M, et al. Autonomous Tissue Retraction in Robotic Assisted Minimally Invasive Surgery – A Feasibility Study. IEEE Robotics and Automation Letters. 2020;5(4):6528-6535. doi:10.1109/LRA.2020.3013914
\bibitem{bendikas_learning_2023} Bendikas R, Modugno V, Kanoulas D, Vasconcelos F, Stoyanov D. Learning Needle Pick-and-Place Without Expert Demonstrations. IEEE Robotics and Automation Letters. 2023;8(6):3326-3333. doi:10.1109/LRA.2023.3266720
4. Cabrelli LC, Pelissari PIBGB, Deana AM, Carneiro AAO, Pavan TZ. Stable phantom materials for ultrasound and optical imaging. Phys Med Biol. 2016;62(2):432. doi:10.1088/1361-6560/62/2/432
\bibitem{dettorre_learning_2022} D’Ettorre C, Zirino S, Dei NN, Stilli A, De Momi E, Stoyanov D. Learning intraoperative organ manipulation with context-based reinforcement learning. Int J CARS. 2022;17(8):1419-1427. doi:10.1007/s11548-022-02630-2
\bibitem{fontanelli_v-rep_2018} Fontanelli GA, Selvaggio M, Ferro M, Ficuciello F, Vendittelli M, Siciliano B. A V-REP Simulator for the da Vinci Research Kit Robotic Platform. In: 2018 7th IEEE International Conference on Biomedical Robotics and Biomechatronics (Biorob). ; 2018:1056-1061. doi:10.1109/BIOROB.2018.8487187
\bibitem{fuller_high_2022} Fuller DB, Crabtree T, Kane BL, et al. High Dose “HDR-Like” Prostate SBRT: PSA 10-Year Results From a Mature, Multi-Institutional Clinical Trial. Frontiers in Oncology. 2022;12. Accessed July 14, 2023. https://www.frontiersin.org/articles/10.3389/fonc.2022.935310
\bibitem{hernandez-quintanar_discovering_2018} Hernandez-Quintanar L, Rodriguez-Salvador M. Discovering new 3D bioprinting applications: Analyzing the case of optical tissue phantoms. Int J Bioprint. 2018;5(1):178. doi:10.18063/IJB.v5i1.178
\bibitem{kazanzides_open_source_2014} Kazanzides P, Chen Z, Deguet A, Fischer GS, Taylor RH, DiMaio SP. An open-source research kit for the da Vinci® Surgical System. In: 2014 IEEE International Conference on Robotics and Automation (ICRA). ; 2014:6434-6439. doi:10.1109/ICRA.2014.6907809
\bibitem{liow_think_2017} Liow MHL, Chin PL, Pang HN, Tay DKJ, Yeo SJ. THINK surgical TSolution-One® (Robodoc) total knee arthroplasty. SICOT-J. 2017;3:63. doi:10.1051/sicotj/2017052
\bibitem{murali_learning_2015}  Murali A, Sen S, Kehoe B, et al. Learning by observation for surgical subtasks: Multilateral cutting of 3D viscoelastic and 2D Orthotropic Tissue Phantoms. In: 2015 IEEE International Conference on Robotics and Automation (ICRA). ; 2015:1202-1209. doi:10.1109/ICRA.2015.7139344
\bibitem{saeidi_autonomous_2022} Saeidi H, Opfermann JD, Kam M, et al. Autonomous robotic laparoscopic surgery for intestinal anastomosis. Science Robotics. 2022;7(62):eabj2908. doi:10.1126/scirobotics.abj2908
\bibitem{shin_autonomous_2019}  Shin C, Ferguson PW, Pedram SA, Ma J, Dutson EP, Rosen J. Autonomous Tissue Manipulation via Surgical Robot Using Learning Based Model Predictive Control. In: 2019 International Conference on Robotics and Automation (ICRA). ; 2019:3875-3881. doi:10.1109/ICRA.2019.8794159


\bibitem{tree_intensity-modulated_2022} Tree AC, Ostler P, Voet H van der, et al. Intensity-modulated radiotherapy versus stereotactic body radiotherapy for prostate cancer (PACE-B): 2-year toxicity results from an open-label, randomised, phase 3, non-inferiority trial. The Lancet Oncology. 2022;23(10):1308-1320. doi:10.1016/S1470-2045(22)00517-4
\bibitem{varier_ambf-rl_2022} Varier VM, Rajamani DK, Tavakkolmoghaddam F, Munawar A, Fischer GS. AMBF-RL: A real-time simulation based Reinforcement Learning toolkit for Medical Robotics. In: 2022 International Symposium on Medical Robotics (ISMR). ; 2022:1-8. doi:10.1109/ISMR48347.2022.9807609

\bibitem{ddpg}  Lillicrap TP, Hunt JJ, Pritzel A, et al. Continuous control with deep reinforcement learning. 2019. arXiv:1509.02971 [cs, stat]
\bibitem{reddy_sqil_2020}  Reddy S, Dragan AD, Levine S. SQIL: IMITATION LEARNING VIA REINFORCEMENT LEARNING WITH SPARSE REWARDS. 2020.
\bibitem{ddpgbc}  Nair A, McGrew B, Andrychowicz M, Zaremba W, Abbeel P. Overcoming Exploration in Reinforcement Learning with Demonstrations. 2018. arXiv:1709.10089 [cs]
\bibitem{col} Goecks VG, Gremillion GM, Lawhern VJ, Valasek J, Waytowich NR. Integrating Behavior Cloning and Reinforcement Learning for Improved Performance in Dense and Sparse Reward Environments. 2020. arXiv:1910.04281 [cs, stat]

\end{thebibliography}
\end{document}